\documentclass[letterpaper]{article} 
\usepackage{aaai25}  
\usepackage{times}  
\usepackage{helvet}  
\usepackage{courier}  
\usepackage[hyphens]{url}  
\usepackage{graphicx} 
\usepackage{natbib}  
\usepackage{caption} 
\usepackage{algorithm}
\usepackage{algorithmic}
\usepackage{amsmath}
\usepackage{booktabs}
\usepackage{array}
\usepackage{amsfonts}
\usepackage{newfloat}
\usepackage{listings}
\DeclareCaptionStyle{ruled}{labelfont=normalfont,labelsep=colon,strut=off} 
\floatstyle{ruled}
\newfloat{listing}{tb}{lst}{}
\floatname{listing}{Listing}
\title{Leveraging Consistent Spatio-Temporal Correspondence \\
for Robust Visual Odometry}
\author{
    Zhaoxing Zhang\textsuperscript{\rm 1}\equalcontrib,
    Junda Cheng\textsuperscript{\rm 1}\equalcontrib,
    Gangwei Xu\textsuperscript{\rm 1},
    Xiaoxiang Wang\textsuperscript{\rm 1},
    Can Zhang\textsuperscript{\rm 1},
    Xin Yang\textsuperscript{\rm 1,2}\thanks{Corresponding author.}
}
\affiliations{
    \textsuperscript{\rm 1}School of Electronic Information and Communications, Huazhong University of Science and Technology\\
    \textsuperscript{\rm 2}Hubei Key Laboratory of Smart Internet, Huazhong University of Science and Technology\\
    \{zzx, jundacheng, gwxu, xiaoxiang0012,can\textunderscore zhang xinyang2014\}@hust.edu.cn

}

\usepackage{bibentry}

\begin{document}
\maketitle

\begin{abstract}
Recent approaches to VO have significantly improved performance by using deep networks to predict optical flow between video frames. However, existing methods still suffer from noisy and inconsistent flow matching, making it difficult to handle challenging scenarios and long-sequence estimation.
To overcome these challenges, we introduce \textbf{S}patio-\textbf{T}emporal \textbf{V}isual \textbf{O}dometry (\textbf{STVO}), a novel deep network architecture that effectively leverages inherent spatio-temporal cues to enhance the accuracy and consistency of multi-frame flow matching.  With more accurate and consistent flow matching, STVO can achieve better pose estimation through the bundle adjustment (BA).
Specifically, STVO introduces two innovative components: 1) the Temporal Propagation Module that utilizes multi-frame information to extract and propagate temporal cues across adjacent frames, maintaining temporal consistency; 2) the Spatial Activation Module that utilizes geometric priors from the depth maps to enhance spatial consistency while filtering out excessive noise and incorrect matches.
Our STVO achieves state-of-the-art performance on TUM-RGBD, EuRoc MAV, ETH3D and KITTI Odometry benchmarks. Notably, it improves accuracy by 77.8\% on ETH3D benchmark and 38.9\% on KITTI Odometry benchmark over the previous best methods.
\end{abstract}



\section{Introduction}
Visual Odometry (VO) is a pivotal technology for estimating a robot's position and orientation by analyzing data from visual sensors. 
Classical  VO methods utilize optimization techniques to either minimize the photometric error of pixel intensities \cite{svo,dso} or minimize the reprojection error of correspondences \cite{orb-slam,orb-slam2,orb-slam3,SR-LIO,SR-LIVO,SDV-LOAM} to estimate trajectories. However, due to limited matching capabilities, classical methods frequently encounter issues with robustness, particularly when the image lacks distinct feature points or the assumption of photometric consistency is not met. In the last decade, there has been a shift towards employing end-to-end deep learning methods \cite{deepvo,deepv2d,ba-net,tartanvo}, which map visual inputs directly to poses. 
End-to-end deep learning methods can achieve more stable matching, but neural networks often struggle to predict 6-degree-of-freedom poses from high-dimensional features directly. As a result, their accuracy often falls short compared to classical optimization backends which have strong geometric constraints.

\begin{figure}[t]
\centering
\includegraphics[width=0.47\textwidth]
{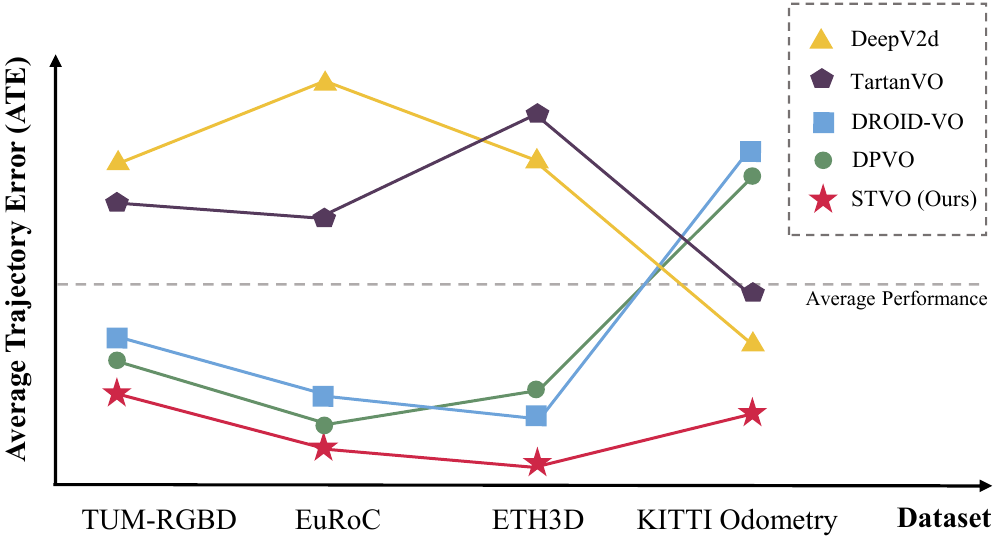} 
\caption{Comparison of STVO with other influential visual odometry methods. Our STVO, denoted by red stars, achieves state-of-the-art performance in all benchmarks.}
\label{resultcompare}
\end{figure}

To address these limitations, a hybrid VO framework \cite{droid,dpvo,salient} has emerged, integrating the strengths of classical methods with deep learning techniques. These frameworks initially employ neural networks to predict flow correspondences and subsequently use correspondences for pose estimation through geometric optimization. By combining the powerful matching capabilities of deep learning with classical geometric constraints, hybrid VO frameworks significantly reduce overfitting and enhance robustness and accuracy.
However, since correspondence estimation is pivotal to the robustness of hybrid VO systems, the persistent issues of noisy and inconsistent flow matching in existing hybrid methods result in significant performance degradation, particularly in challenging scenarios. Furthermore, VO systems commonly experience cumulative trajectory drift, which is exacerbated by these matching problems, resulting in increasingly inaccurate estimations over long sequences.

\begin{figure}[t]
\centering
\includegraphics[width=0.47\textwidth]
{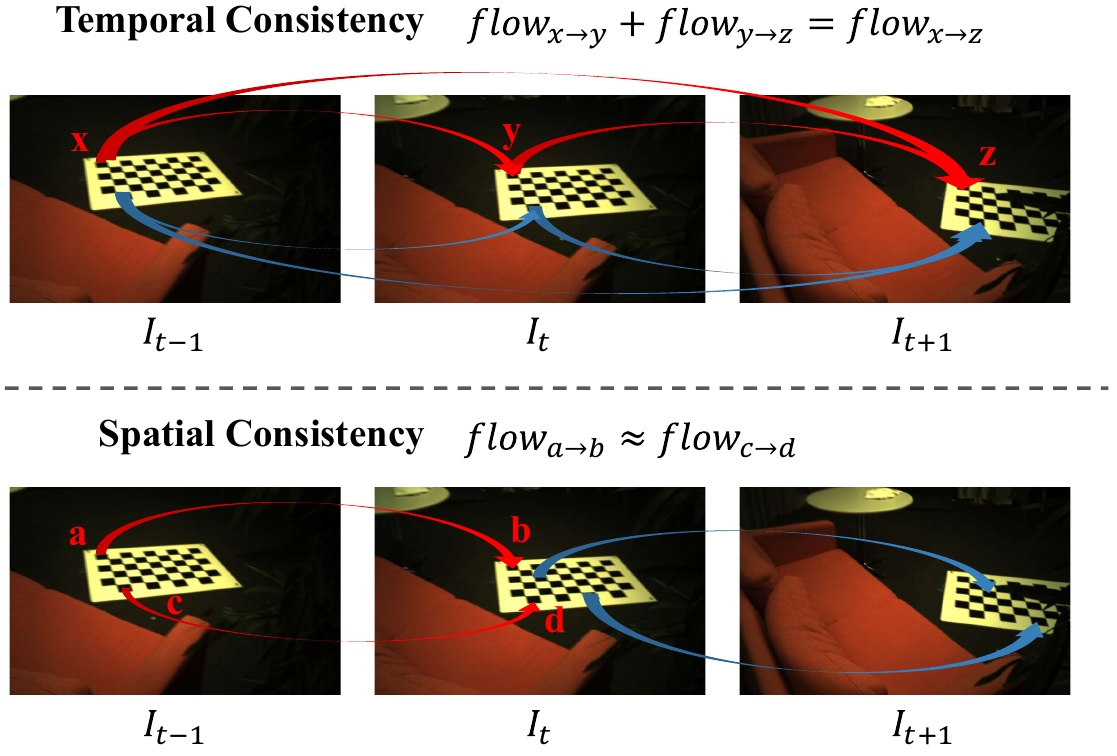} 
\caption{Diagram of Temporal Consistency and Spatial Consistency Across Multiple Frames}
\label{spatialtemporal}
\end{figure}

Existing hybrid methods rely solely on the information between two frames to obtain correspondences. However, flow estimated from just two frames is prone to noise and inconsistencies, especially in challenging regions. These methods overlook the potential of VO systems to jointly estimate poses across multiple frames in local windows, which could also be leveraged for multi-frame joint optical flow estimation. This approach can significantly enhance the accuracy and robustness of optical flow by optimally integrating spatio-temporal cues. With more consistent flow matching in both spatial and temporal domains, VO can achieve better pose estimation through the bundle adjustment (BA) step.
This insight is driven by two key observations, as illustrated in Figure \ref{spatialtemporal}. First, there is temporal consistency across multiple frames, where adjacent frames share similar motion trends and are constrained by flow consistency. Second, each frame exhibits spatial consistency, where different points on the same object maintain highly uniform motion patterns.

Building on these observations, we introduce Spatio-Temporal Visual Odometry (STVO), a novel deep network architecture that integrates spatio-temporal cues to optimize multi-frame optical flow matching. 
Firstly, we introduce a Temporal Propagation Module which leverages multi-frame information to extract and propagate temporal cues across adjacent frames. 
The Temporal Propagation Module maintains a motion state for each source frame and iteratively uses currently predicted optical flows to warp motion state across all adjacent frames, enabling the accurate reflection of motion dynamics. Finally, the temporal information obtained from adjacent frames is propagated back to the source frame and updates the motion state for the next iteration.
Secondly, we introduce the Spatial Activation Module. This module innovatively utilizes depth information and geometric priors to enhance spatial consistency in flow estimation. It leverages geometric information from depth maps to create a Spatial Attention Matrix, which employs a attention-based approach to model spatial cues. The Spatial Attention Matrix is then used to activate the coarse context and correlation features, enabling comprehensive spatial understanding and filtering out noise and incorrect matches.

Extensive experiments demonstrate that our STVO outperforms all prior works across all four real-world benchmarks. Specifically, on the challenging ETH3D dataset, STVO shows a 77.8\% improvement over the previous best method. Additionally, on the long-sequence KITTI Odometry benchmark, it achieves a 38.9\% improvement over the previous best method.

In summary, our main contributions are as follows:
\begin{itemize}
\item To the best of our knowledge, we are the first to highlight the significance of both spatial and temporal consistency for matching in Visual Odometry. We introduce Spatio-Temporal Visual Odometry (STVO), a novel deep network architecture that harnesses spatio-temporal cues to mitigate performance degradation in challenging scenarios and long-sequence estimations.
\item We propose the Temporal Propagation Module, which capitalizes on the high correlation between neighboring frames to enhance temporal consistency.
\item We propose the Spatial Activation Module, which utilizes depth information and geometric priors to maintain spatial consistency and filter out noise and incorrect matches.
\item Our STVO achieves state-of-the-art accuracy on four benchmarks: TUM-RGBD, EuRoC MAV, ETH3D, and KITTI Odometry, while also showcasing outstanding performance under challenging conditions and excellent resistance to drift in long sequences.
\end{itemize}

\begin{figure*}[t]
\centering
\includegraphics[width=0.97\textwidth]{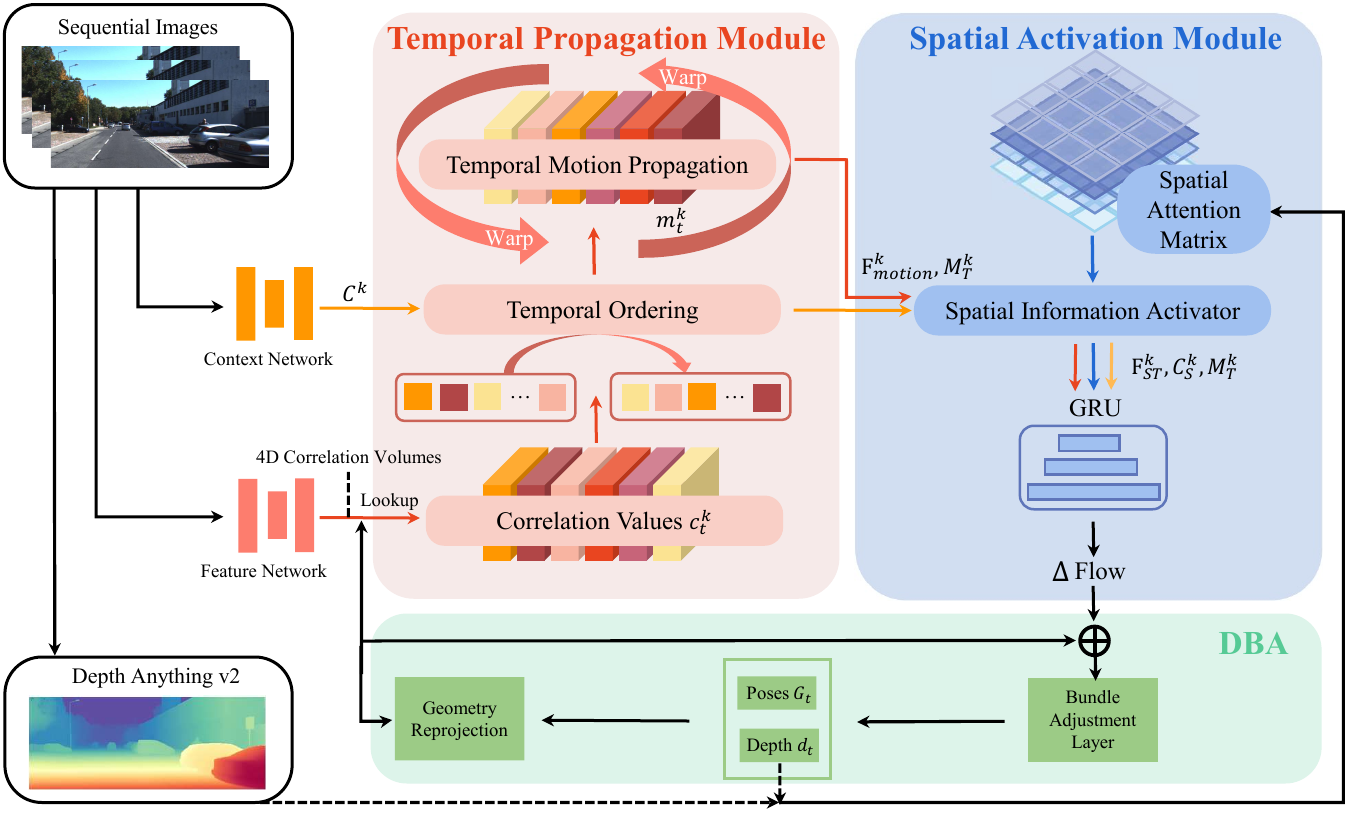} 
\caption{Overview of STVO. The architecture consists of three key modules: 1) Temporal Propagation Module, which enhances temporal consistency; 2) Spatial Activation Module, which maintains spatial consistency and filters out incorrect matches; 3) Differentiate Bundle Adjustment (DBA) Module, which updates poses and depths using optical flow estimates. The dashed lines in STVO indicate that the depth map input can be flexibly chosen between the depth generated by Depth Anything V2 and the output depth map of Bundle Adjustment. Both input options have demonstrated significant effectiveness.}
\label{key_figure}
\end{figure*}
\section{Related Work}
Visual odometry methods can be broadly categorized into classical methods, deep learning-based methods and Hybrid VO framework.

\subsubsection{Classical Methods} can be further divided into direct and indirect approaches. \textit{Direct} approaches operate based on the photometric constancy assumption, obtain pose by minimizing the photometric error of pixel intensities across the images \cite{svo,dso,LSD-SLAM}. \textit{Indirect} approaches use hand-crafted features to find correspondences and then minimize the reprojection error on the correspondences \cite{orb-slam,orb-slam2,orb-slam3}. A common issue with classical methods is that handcrafted features often lack robustness, making VO systems prone to failure.

\subsubsection{Deep Learning-Based Methods} have been proposed to enhance robustness. Early methods replaced traditional handcrafted feature extraction with neural networks for feature extraction \cite{SuperPoint,LF-Net} or adopted end-to-end approaches to predict the pose \cite{deepvo,deepv2d,tartanvo}. Although learning-based VO has shown to improve accuracy and robustness compared to classical approaches, they still face the challenge of significant performance degradation when dealing with data that differs from training distributions \cite{DF-VO,zhangdong_causal,zhangdong_feature,wxx,coatrsnet,adaptive}.

\subsubsection{Hybrid VO Framework} was proposed to combine the strengths of classical methods and deep learning methods. There has already been a series of outstanding works in this field such as 
\cite{droid,dpvo,salient,variance,romeo,devo}, among them, two particularly notable and closely related to our work are DROID-SLAM and DPVO. DROID-SLAM integrates RAFT for iterative dense optical flow prediction to obtain a more robust correspondence, which greatly improves the robustness of VO system. To improve the efficiency of VO, DPVO \cite{dpvo} replaces dense optical flow tracking with patch-based tracking. However, this patch-based matching approach can exacerbate instability issues in extreme scenarios, such as motion blur or highly dynamic scenes. 

\subsubsection{Optical Flow} is the task of estimating dense 2D pixel-level motion between a pair of frames. 
Many deep learning-based optical flow methods \cite{spynet,pwcnet,GMA,KPA,videoflow,xu2022attention,xu2023iterative,xu2023accurate,xu2024igev++} have shown outstanding performance.
A recent standout work, RAFT \cite{raft}, employs a multi-scale search and recurrent manner to estimate flow, effectively balancing both accuracy and efficiency. Many hybrid VO systems use RAFT-style networks to obtain optical flow, thereby improving VO performance. However, these systems typically use RAFT merely as a module for obtaining two-frame correspondences, without fully integrating the characteristics of VO. In fact, the local sliding window in VO contains rich temporal and motion information, which can be effectively utilized to enhance the robustness of optical flow estimation, rather than relying on isolated pairwise two-frame matching. This is precisely what STVO aims to achieve.

\section{Background}
We choose DROID-VO as our baseline (the VO front-end of DROID-SLAM \cite{droid}).
In this section, we provide a brief overview of the relevant components of DROID-VO to help readers better understand our approach.

\textbf{Representation.}
Our network operates on an ordered collection of images $\left\{  {{I}_{t}} \right\}^{N}_{t=0}$, camera poses ${G}_{t} \in SE(3)$, and inverse depths ${d}_{t}\in {\mathbb{R}}^{H\times W}_{+}$ follow DROID-VO.
We adopt a frame-graph $(\mathcal{V}, \mathcal{E})$ to represent co-visibility between keyframes. And the frame-graph is built dynamically during training and inference to ensure that, within the local window, only the $r$ reference frames closest to the target keyframe are retained.

\textbf{Update Operator.} 
The key to DROID-VO's outstanding performance lies in using the RAFT-based update operator to iteratively refine the optical flow. This approach projects the estimates of depth ${d}_{t}$ and pose ${G}_{t}$ in each Bundle Adjustment (BA) iteration to obtain initial optical flow ${f}^{k}_{t}$ and uses the GRU module to refine the flow. To be specific, the correlation features and context features are injected into the GRU, which maintains a hidden state $h^k $ during the update process and outputs the revision flow field ${r}_{ij}\in {\mathbb{R}}^{H\times W\times 2}$ and the associated confidence map ${w}_{ij}\in {\mathbb{R}}^{H\times W\times 2}_{+}$. The improved optical flow is then used to optimize more accurate depth and pose in the BA module, thereby forming a feedback loop that enhances the overall accuracy.

\textbf{Differentiable Bundle Adjustment.}
After obtaining the predicted flow revisions and the flow confidence, the differentiate bundle adjustment (DBA) Layer is applied to get the updated poses and pixel-wise depths. The DBA objective is as follows:
\begin{equation}
{T}^{\ast },{d}^{\ast }=\arg \min _{T, d}\sum_{(i, j) \in \mathcal{E}}\left\|\hat{P}_{i j}-\tilde{P}_{i j}\right\|_{\Sigma_{i j}}^2 
\end{equation}

where $\hat{P}_{ij}$, $\tilde{P}_{ij}$ and $\Sigma_{ij}$ denote the reprojected position, the estimated flow, and the confidence weights from the $I_i$ to the $I_j$, respectively.

\section{Method}

\subsection{Overview}
Figure \ref{key_figure} presents the overall architecture of our network. 
STVO consists of four main steps: 1) \emph{Feature Extraction Network}, which extracts features used to compute the optical flow cost volume and generates correlation values ${c}^{k}_{t}$. 2) \emph{Temporal Propagation Module}, by imposing temporal consistency constraints on the correlation values, a more robust matching cost is obtained. 3) \emph{Spatial Activation Module}, which applied the spatial consistency constraints to the correlation values to derive the final matching cost volume, then through a GRU module to obtain the revision flow. 4) \emph{Differentiable Bundle Adjustment (DBA) Module}, the refined optical flow is utilized by the DBA module to optimize the depth and pose. The optimized depth ${d}_{t}$ and pose ${G}_{t}$ are then used for geometric reprojection, providing an initial optical flow estimate for the next iteration, thereby creating a positive feedback loop. The key contributions of STVO lie in two main components: the Temporal Propagation Module and the Spatial Activation Module, which can complement each other to achieve a more stable optical flow, further enhancing the potential of the positive feedback loop and improving the accuracy of pose estimation.
\subsection{Temporal Propagation Module}
\begin{figure}[h]
\centering
\includegraphics[width=0.46\textwidth]
{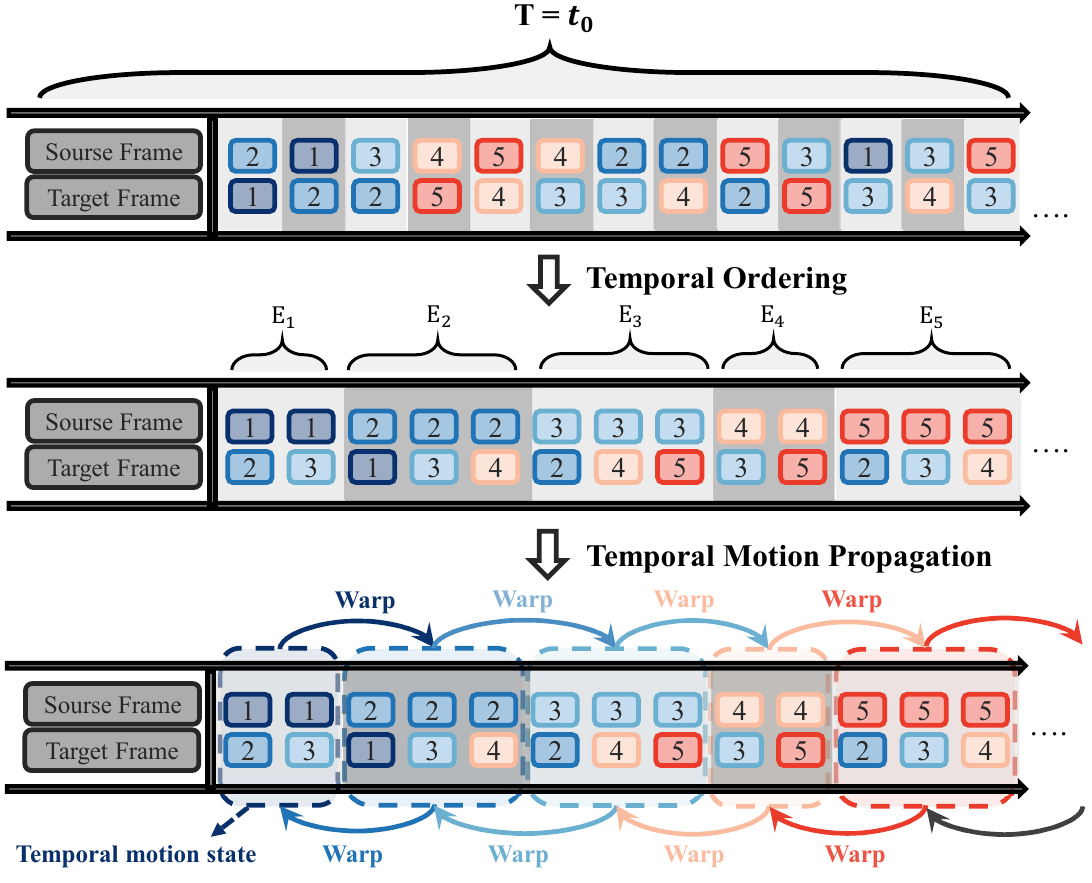} 
\caption{Diagram of Temporal Propagation Module.}
\vspace{-7pt}
\label{Temporal_motion}
\end{figure}

\subsubsection{Temporal Ordering.} 
As shown in the first row of Figure 3, the dynamic addition and removal of keyframes within the local window causes the edges $\mathcal{E}$ in the frame graph to lose their chronological order, thereby disrupting the temporal relationships between frames. This disruption makes it challenging to leverage temporal consistency for constraining the optical flow. To address this, we applied temporal ordering to the correlation values, sorting them according to the sequence of their source frames. 
For simplicity, we denote the set of edges in the frame-graph that originate from the same source frame $t$ as the Source Edge Set $E_t$ in the following text, where \( E_t = \{ (t, j) \mid t \in \mathcal{V}\) and \((t, j) \in \mathcal{E}\}\).

\begin{table*}[h]
  \centering
  \begin{tabular}{lccccccccccc}
    \toprule
    & 360 & desk & desk2 & floor & plant & room & rpy & teddy & xyz & Avg\\
    \midrule
    ORB-SLAM3 & - & \textbf{0.017} & 0.210 & - & 0.034 & - & - & - & \textbf{0.009} & - \\
    DSO & 0.173 & 0.567 & 0.916 & 0.080 & 0.121 & 0.379 & 0.058 & - & 0.036 & - \\
    \midrule
    DeepV2d & \textbf{0.144} & 0.105 & 0.321 & 0.628 & 0.217 & \textbf{0.215} & 0.046 & 0.294 & 0.051 & 0.225 \\
    TartanVO & 0.178 & 0.125 & 0.122 & 0.349 & 0.297 & 0.333 & 0.049 & 0.339 & 0.062 & 0.206\\
    DROID-VO & \underline{0.161} & \underline{0.028} & 0.099 & \textbf{0.033} & \underline{0.028} & 0.327 & \underline{0.028} & 0.169 & 0.013 & 0.098\\
    DPVO & 0.165 & 0.034 & \textbf{0.042} & 0.050 & 0.036 & 0.388 & 0.034 & \textbf{0.057} & \underline{0.012} & \underline{0.091} \\
    STVO(Ours) & 0.171 & 0.031 & \underline{0.068} & \underline{0.035} & \textbf{0.027} & \underline{0.242} & \textbf{0.027} & \underline{0.111} & 0.015 & \textbf{0.080} \\
    \bottomrule
  \end{tabular}
\caption{ Performance comparisons on the TUM-RGBD Dataset on ATE[m]. 
$(-)$ indicates that the method failed to track. We use \textbf{bold} and \underline{\hspace{0.25cm}} to highlight the methods that rank 1st and 2nd.}
\label{table_tum}
\end{table*}
\subsubsection{Temporal Motion Propagation.} To leverage the benefits of temporal ordering, allowing the system to utilize the sequential information and enhance temporal consistency, STVO maintains a motion state ${m}^{k}_{t}\in {R}^{H\times W\times {D}_{m}}$ for each Source Edge Set $E_t$, where ${m}^{0}_{t}$ is randomly initialized. The motion state ${m}^{k}_{t}$ contains motion information at the $k$-th iteration for $I_t$ and will be updated in each iteration. 

In iteration $k$, for each edge $(m,n)$ in the Source Edge Set $E_m$, the motion state ${m}^{k}_{m}$ is warped using the predicted optical flow $f_{m\rightarrow n}$.
This warping produces the corresponding dynamic motion state ${M}^{k}_{m \rightarrow n}\in{R}^{H\times W\times {D}_{m}}$, enabling the target frame to leverage the motion information from the source frame. We then concatenate the source frame's motion state ${m}^{k}_{m}$, the target frame's motion state ${m}^{k}_{n}$, and the acquired dynamic motion state ${M}^{k}_{m \rightarrow n}$ to obtain the temporal motion state ${M}^{k}_{T}\in{R}^{H\times W\times 3{D}_{m}}$ as below, which providing a comprehensive representation of the temporal motion across this frame edge $(m,n)$.
\begin{equation}
\begin{aligned}
{M}^{k}_{m \rightarrow n}=\operatorname{Warp}({m}^{k}_{m};{f}^{k}_{m \rightarrow n}),
\\
{M}^{k}_{T}=\operatorname{Concat}({m}^{k}_{m},{M}^{k}_{m \rightarrow n},{m}^{k}_{n})
\label{eq2}
\end{aligned}
\end{equation}

The we concatenate the correlation feature ${F}^{k}_{corr}$ and our temporal motion state ${M}^{k}_{T}$ to generate the temporal motion feature ${F}^{k}_{motion}\in{R}^{H\times W\times {D}_{M}}$ and local motion state ${m}^{k+1}_{m \rightarrow n}$ by \emph{TemporalEncoder}, which is a cascaded two-layer 2D convolution. Subsequently, STVO propagates all the collected local motion states back to $I_m$ to update the motion state for the next iteration.
\begin{equation}
\begin{gathered}
{F}^{k}_{motion}, {m}^{k+1}_{m \rightarrow n} = \operatorname{TemporalEncoder}({F}^{k}_{corr},{M}^{k}_{T}), \\
m^{k+1}_{m} = \frac{1}{|T_m|} \sum_{n \in T_m} m^{k+1}_{m \rightarrow n}\\
\end{gathered}
\end{equation}
where \( T_m = \{ n \mid (m, n) \in E_m \} \) represents the collection of all target frames for each frame image $I_m$, and $|T_m|$ is the number of frame edges in $E_m$. The temporal motion feature ${F}^{k}_{motion}$ and the temporal motion state ${M}^{k}_{T}$ are then transferred to the Spatial Activation Module to further enhance spatial consistency. 

\begin{figure}[h]
\centering
\includegraphics[width=0.47\textwidth]
{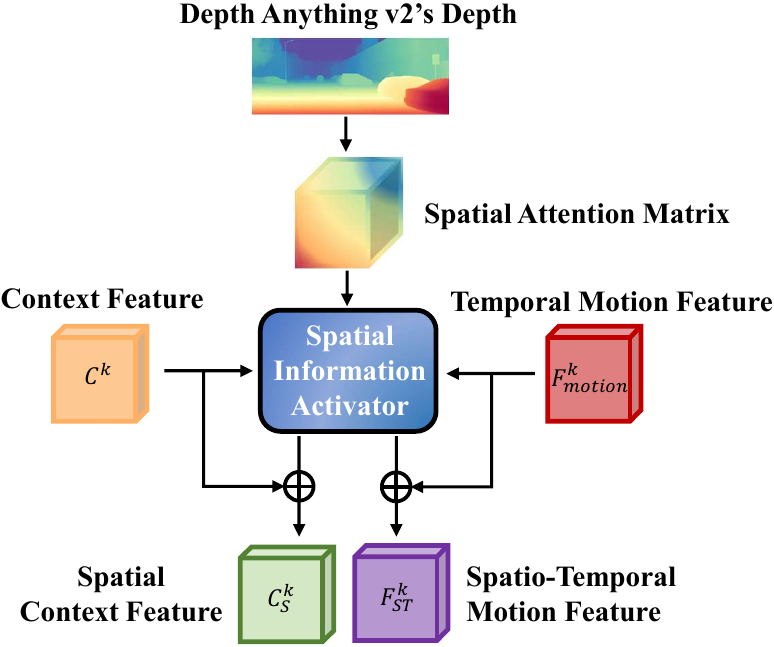} 
\caption{Diagram of Spatial Activation Module}
\label{Spatial_Activation}
\end{figure}

\subsection{Spatial Activation Module}


\subsubsection{Spatial Attention Matrix.}
As shown in Figure \ref{key_figure}, for each image $I_m\in {R}^{H_1\times W_1\times 3}$, we generate a corresponding depth map $D_m\in {R}^{H \times W}$ by using the Depth Anything v2 small model or using our BA depth (output of DBA module), where $H = H_1/8$, $W = W_1/8$. 
To enhance flow consistency, we exploit the spatial consistency information embedded in the depth map through spatial attention. Specifically, the query feature $q_m$ and key feature $k_m$ are derived by projecting the depth map $D_m$ into a higher-dimensional feature space. These features are used to compute the Spatial Attention Matrix (SAM), which captures the spatial relationships and context dependencies within the depth map $D_m$:
\begin{equation}
\begin{gathered}
q_m = D_m W_q, k_m = D_m W_k \\
SAM\, =\, \sigma\left(q_m k_m^{\top}\right)
\end{gathered}
\end{equation}
where $W_q$,$W_k\in {R}^{1 \times D_{in}}$ are projection matrices, and $q_m,k_m\in {R}^{H W \times D_{in}}$ denotes query, key features. $\sigma(\cdot)$ denotes the softmax function.


\subsubsection{Spatial Information Activator.} 
As shown in Figure \ref{Spatial_Activation}, the Spatial Attention Matrix is utilized in the Spatial Information Activator to aggregate object consistency information and filter out irrelevant information, thereby making the matching features more robust. Specifically, we apply attention aggregation to both the temporal motion feature ${F}^{k}_{motion}$ and context feature ${C}^{k}$, resulting in our spatio-temporal motion Feature  ${F}^{k}_{ST}$ and spatial context feature ${C}^{k}_{S}$ respectively as below:

\begin{equation}
\begin{gathered}
{C}^{k}_{S} = {C}^{k} + \alpha_c \cdot SAM\cdot {C}^{k} \\
{F}^{k}_{ST} = {F}^{k}_{motion} + \alpha_f \cdot SAM\cdot {F}^{k}_{motion} 
\end{gathered}
\end{equation}
where $\alpha_c$ and $\alpha_f$ are learned scalar parameters. Once the matching features have been optimized through our Temporal Propagation Module and Spatial Activation Module, they are combined with the temporal motion state ${M}^{k}_{T}$ and fed into the GRU module to obtain the refined optical flow for the current iteration.

\setlength{\tabcolsep}{1mm}  
\begin{table*}[h]
  \centering
  \begin{tabular}{lccccccccc}
    \toprule
    & cables & camera\underline{~}shake & ceiling & desk\underline{~}changing & einstein & mannequin & plant\underline{~}scene & sfm\underline{~}lab\underline{~}room & Avg\\
    \midrule
    ORB-SLAM3 & 0.210 & - & - & - & 0.138 & - & - & - & - \\
    DSO & 0.174 & - & - & 1.406 & 0.301 & 0.805 & - & 0.827 & -\\
    \midrule
    DeepV2d & 0.229 & 0.100 & 1.974 & 0.964 & 0.069 & 0.559 & 0.262 & 0.057 & 0.526\\
    TartanVO & 0.364 & 0.107 & 1.889 & 0.851 & 0.120 & 0.227 & 0.651 & 0.277 & 0.561\\
    DROID-VO & 0.030 & \textbf{0.034} & \underline{0.193} & \underline{0.186} & \textbf{0.002} & \textbf{0.009} & \textbf{0.005} & 1.066 & \underline{0.190}\\
    DPVO & \textbf{0.020} & 0.071 & 0.398 & 1.405 & 0.007 & 0.020 & 0.017 & \underline{0.060} & 0.249\\
    STVO(Ours) & \underline{0.028} & \underline{0.038} & \textbf{0.183} & \textbf{0.045} & \underline{0.003} & \underline{0.013} & \underline{0.009} & \textbf{0.015} & \textbf{0.042}\\
    \bottomrule
  \end{tabular}
\caption{ Performance comparisons on the ETH3D Dataset on ATE[m]. $(-)$ indicates that the method failed to track. }
\label{table_ETH3D}
\end{table*}

\begin{table*}[h]
  \centering
  \begin{tabular}{lccccccccccc}
    \toprule
    & 00 & 03 & 04 & 05 & 06 & 07 & 08 & 09 & 10 & Avg\\
    \midrule
    ORB-SLAM3 & 74.69 & \textbf{0.64} & 1.81 & 33.22 & 47.07 & 16.20 & 54.46 & 46.61 & \textbf{7.13} & 31.31 \\
    DSO & \underline{48.04} & \underline{0.80} & \textbf{0.36} & 48.45 & 57.59 & 53.67 & 113.02 & 92.19 & 11.03 & 47.23 \\
    \midrule
    DeepV2d & 101.65 & 7.15 & 4.08 & \underline{27.05} & \textbf{7.39} & \underline{8.70} & \textbf{18.91} & \textbf{10.13} & 14.77 & \underline{22.20} \\
    TartanVO & 63.84 & 7.71 & 2.89 & 54.61 & 24.67 & 19.29 & 59.55 & 32.61 & 25.04 & 32.25\\
    DROID-VO & 109.00 & 5.57 & \underline{1.05} &  60.37 & 38.03 & 21.41 & 105.64 & 73.04 & 13.77& 47.53\\
    DPVO & 108.88 & 1.60 & 1.55 & 56.60 & 59.37 & 17.64 & 97.48 & 62.94 & \underline{10.03} & 46.23 \\
    STVO(Ours) & \textbf{16.13} & 2.15 & 1.93 & \textbf{19.49} & \underline{14.07} & \textbf{7.72} & \underline{29.18} & \underline{17.28} & 14.16 & \textbf{13.56}  \\
    \bottomrule
  \end{tabular}
\caption{ Performance comparisons on the KITTI Odometry Dataset on ATE[m]. }
\label{table_KITTI}
\end{table*}

\subsection{Implementation Details}
STVO is implemented using PyTorch \cite{pytorch} and C++. The training strategy for STVO is almost consistent with that of DROID-VO, which is conducted on monocular images from the synthetic TartanAir dataset \cite{tartanair}. We train our network for 250k steps with a batch size of 4, resolution 384 × 512, and 7 frame clips, and unroll 15 update iterations.

\section{Experiments}
In the experimental section, we validated the effectiveness of the STVO design through comprehensive quantitative and qualitative comparisons. Additionally, we performed ablation studies to analyze the contributions of each component.

\subsection{Quantitative Comparison}
In this section, we conduct experiments on popular real-world VO benchmarks to quantitatively evaluate the effectiveness of our approach. 
Following prior works, We assess the estimated trajectory on Absolute Trajectory Error (ATE) using scale alignment with evo \cite{evo}.
We compare STVO with classical methods, such as ORB-SLAM3 \cite{orb-slam3}, DSO \cite{dso}, and influential deep learning methods like DeepV2d \cite{deepv2d}, TartanVO \cite{tartanvo}, DROID-VO \cite{droid}, and DPVO \cite{dpvo} across four public datasets: TUM-RGBD \cite{tum}, EuRoC MAV \cite{euroc}, ETH3D \cite{eth3d} and KITTI Odometry \cite{kitti}. 
For a fair comparison, we present the results for DPVO as the average of five runs, instead of using the median of five runs as reported in \cite{dpvo}. This choice is made because the median method tends to exclude outliers and does not fully represent the model's average performance.

\subsubsection{TUM-RGBD.} The TUM-RGBD dataset \cite{tum} consists of indoor scene data captured using handheld cameras, which introduces challenges like rolling shutter artifacts, motion blur, and significant rotations. In Table \ref{table_tum}, we benchmark on TUM-RGBD and compare to SOTA. 
STVO achieves the best or second-best results in 6 out of 9 sequences and delivers the best overall average performance. Compared to DPVO and DROID-VO, it shows improvements of 12\% and 18\%, respectively. 

\subsubsection{EuRoC MAV.}
In Table \ref{table_EuRoC}, we present our benchmarking results on the EuRoC MAV\cite{euroc} dataset. EuRoC MAV is a widely used benchmark for evaluating VO systems, collected using Micro Aerial Vehicles (MAV). 
The dataset includes fast and dynamic camera movements that pose significant challenges for VO systems.
To align with the practices of similar studies such as DROID-VO and DPVO, we process every other frame, effectively doubling the system's frame rate to 40 FPS. 
STVO outperforms previous methods on the majority of sequences on EuRoC MAV. The average error of STVO is $5\%$ lower than the state-of-the-art DPVO method and $21\%$ lower than DROID-VO.

\begin{table}[h]
\small
  \centering
    \begin{tabular}{lccccccc}
      \toprule
      & V101 & V102 & V103 & V201 & V202 & V203 & Avg \\
      \midrule
      ORB-SLAM3 & \textbf{0.035} & 0.139 & 0.713 & 1.352 & \underline{0.056} & 0.632 & 0.487 \\
      DSO & 0.089 & \underline{0.107} & 0.903 & \textbf{0.044} & 0.132 & 1.152 & 0.404 \\
      \midrule
      DeepV2d & 0.717 & 0.695 & 1.483 & 0.839 & 1.052 & 0.591 & 1.173 \\
      TartanVO & 0.447 & 0.389 & 0.622 & 0.433 & 0.749 & 1.152 & 0.632 \\
      DROID-VO & 0.103 & 0.165 & 0.158 & 0.102 & 0.115 & \textbf{0.204} & 0.141 \\
      DPVO & \underline{0.050} & 0.148 & \underline{0.093} & 0.086 & \textbf{0.049} & 0.282 & \underline{0.118} \\
      STVO(Ours) & 0.055 & \textbf{0.098} & \textbf{0.078} & \underline{0.064} & 0.125 & \underline{0.248} & \textbf{0.111} \\
      \bottomrule
    \end{tabular}%
  \caption{Performance comparisons on EuRoC MAV dataset.}
  \label{table_EuRoC}
\end{table}


\subsubsection{ETH3D.}
In Table \ref{table_ETH3D}, we use the SLAM benchmark of ETH3D \cite{eth3d}. 
It includes eight distinct scenes, each comprising multiple sequences. We consistently select the first sequence from each scene for a fair comparison. 
The ETH3D dataset covers a wide range of scenarios, with many sequences being particularly challenging due to low lighting and significant motion blur. This diversity and difficulty allow for a comprehensive evaluation of VO systems' performance.
For instance, in the desk\_changing sequence, where objects move quickly, both classical methods and the DPVO algorithm, which relies on randomly selecting image patches between only two frames, struggle to achieve stable matching, leading to failure. In contrast, STVO demonstrated the best performance with an error of only 0.045 meters, showcasing its robustness and precision. Notably, our approach also achieved the lowest average error across all test sequences. Compared to DPVO, we reduced the average error from 0.249 meters to 0.042 meters, representing an improvement of $83.1\%$.

\subsubsection{KITTI Odometry.} In Table \ref{table_KITTI}, we benchmark on the KITTI Odometry \cite{kitti} dataset. KITTI Odometry is designed for autonomous driving applications. 
Among existing methods, performance on KITTI Odometry has been generally unsatisfactory. Apart from STVO, DeepV2d demonstrates the best performance due to being the only method trained on the KITTI dataset. However, STVO not only exceeds all other methods, but also achieves a notable $38.9\%$ improvement over DeepV2d, without any fine-tuning on KITTI. This highlights STVO's excellent ability to mitigate trajectory drift in long sequences and its strong generalization capability.

\subsection{Qualitative Comparison}
In this section, we qualitatively compare trajectories and optical flow to validate the effectiveness of STVO.
\begin{figure}[h]
\centering
\includegraphics[width=0.47\textwidth]{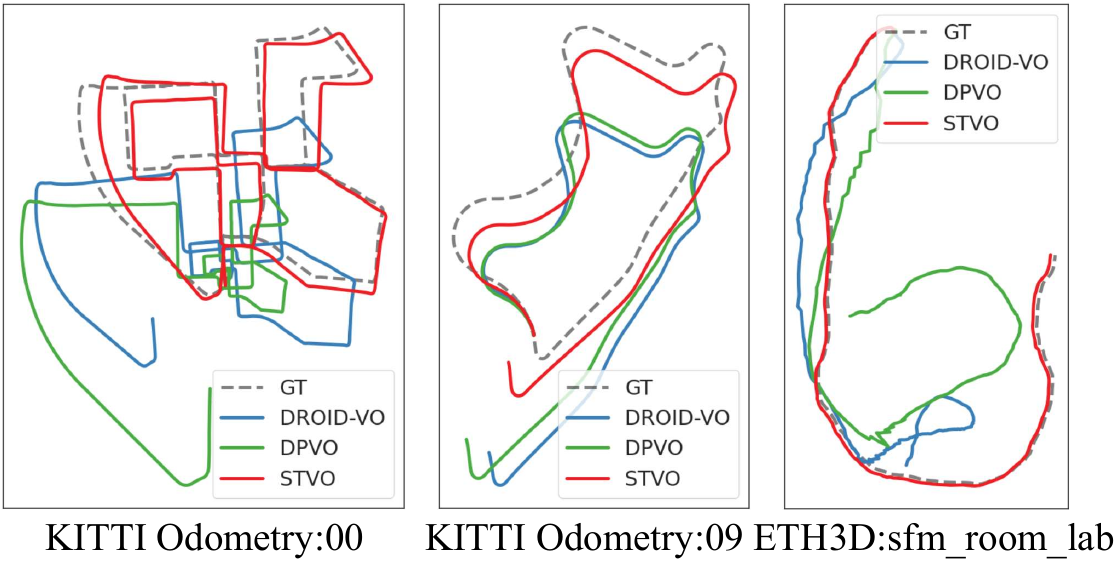} 
\caption{Visualization comparison of trajectory. }
\label{traj_visual}
\end{figure}

\subsubsection{Trajectory Comparison.} As shown in Figure \ref{traj_visual}, both DROID-VO and DPVO exhibit significant trajectory drift in the KITTI Odometry long sequences 00 and 09, which cover distances of 3724m and 1705m, respectively. In contrast, STVO shows remarkable resistance to drift, highlighting its superior performance in long-sequence estimation. For ETH3D sequence sfm\_room\_lab, large area repeated texture increases the difficulty of matching, resulting in serious trajectory drift of DPVO and DROID-VO algorithms that only rely on two frames of information for matching.
However, STVO consistently aligns with the ground truth, demonstrating its ability to maintain accurate trajectory estimation even in challenging situations.

\begin{figure}[h]
\centering
\includegraphics[width=0.47\textwidth]{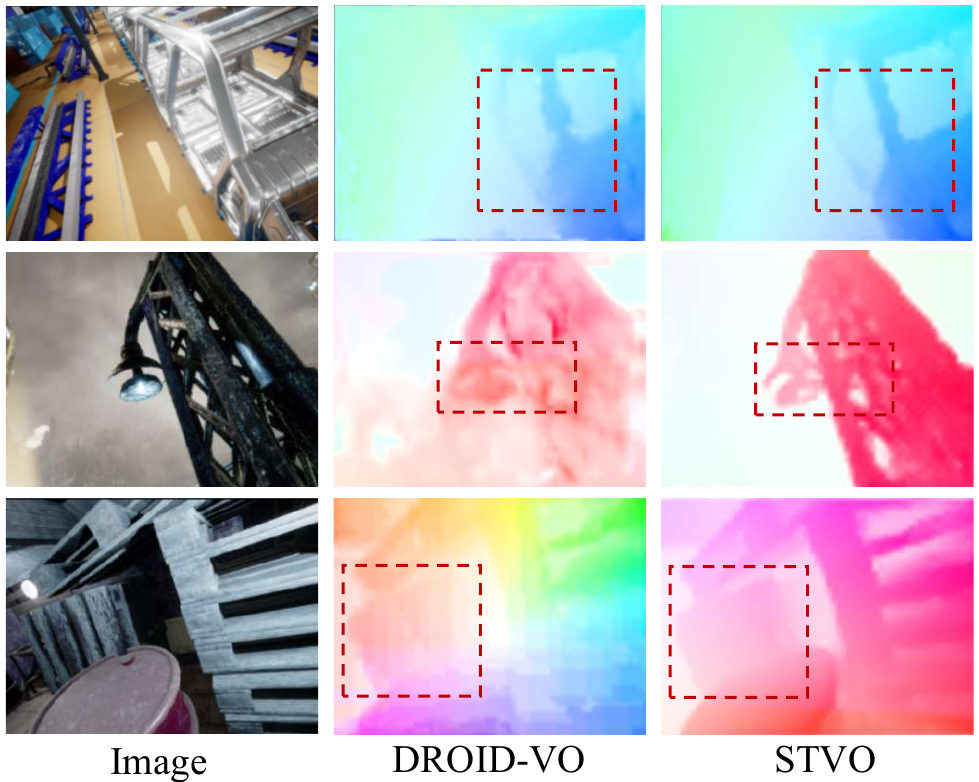} 
\caption{Visual comparison of optical flow in challenging conditions: highlights, fog, and repetitive structures. }
\label{flow}
\end{figure}

\subsubsection{Flow Comparison.} As shown in Figure \ref{flow}, compared to DROID-VO, STVO has a sharper optical flow and a more accurate structure due to the efficient integration of the Spatial Activation Module with spatial information. Even when handling complex situations, STVO still demonstrates superior stability and accuracy in optical flow estimation.

\subsection{Ablations}
We conduct ablation studies in Table \ref{tab:ablation} to validate each component of STVO on the TUM-RGBD dataset by comparing the model's average ATE, GPU memory usage and average frames per second (FPS) under different settings. DROID-VO serves as our baseline. By individually adding our Spatial Activation Module and the Temporal Propagation Module to the baseline (denoted as Base + SAM and Base + TPM), we achieve an improvement in ATE by 9.2$\%$ and 13.3$\%$ respectively without introducing significant memory cost. Moreover, their combined use yields even better results, i.e. our final STVO model. Specifically, to evaluate the impact of Depth Anything V2's geometry priors, we replace Depth Anything V2's depth maps with those from Bundle Adjustment (denoted as DepAny  $\rightarrow$ BA depth). The result shows that using BA depth alone can also significantly improve VO performance with fewer memory costs compared to STVO. This indicates that the SAM's improvement is not solely derived from the robust priors provided by Depth Anything V2, but rather from our design that leverages depth information to enforce motion consistency. 
This highlights the superiority of our Spatial Activation Module. 

\begin{table}[h]
\centering
\begin{tabular}{lcccccc}
\toprule
\textbf{Ablation} & \textbf{SAM} & \textbf{TPM} &  \textbf{ATE} & \textbf{GPU} &  \textbf{FPS}\\
\midrule
DROID-VO (Baseline) &  & & 0.098 &4.2 & 9.8 \\
Base+SAM & \checkmark &  & 0.089 &5.6 &7.8\\
Base+TPM &  & \checkmark & 0.085 &4.8 &9.1\\ 
Full model (STVO) & \checkmark &  \checkmark & 0.080 &6.0 &7.3 \\
DepAny $\rightarrow$ BA depth & \checkmark & \checkmark & 0.082&5.1 &8.0\\
\bottomrule
\end{tabular}
\caption{Ablation Study on the Designs of STVO.}
\label{tab:ablation}
\end{table}






\section{Conclusion}
We present STVO, a novel deep Visual Odometry architecture that integrates  spatio-temporal cues to enhance both spatial and temporal consistency in multi-frame matching, effectively mitigating performance degradation in challenging scenarios and long-sequence estimations.
STVO outperforms all previous methods on TUM-RGBD, EuRoC MAV, ETH3D, and KITTI Odometry.
STVO is the first to highlight the significance of spatial and temporal consistency in multi-frame flow matching for VO.

\clearpage

\section{Acknowledgments}
This work is supported by the National Natural Science Foundation of China (62122029, 62472184), the Fundamental Research Funds for the Central Universities, and the National Natural Science Foundation of China (623B2036).

\bibliography{aaai25}



\end{document}